# High Density Noise Removal By Using Cascading Algorithms

Arabinda Dash
Department of CSE & IT
Veer Surendra Sai University of Technology
Burla, Odisha, India
Email: arabinda006@gmail.com

Sujaya Kumar Sathua
Department of CSE & IT
Veer Surendra Sai University of Technology
Burla, Odisha, India
Email: kumarsujaya@gmail.com

*Abstract*—An advanced non-linear cascading filter algorithm for the removal of high density salt and pepper noise from the digital images is proposed. The proposed method consists of two stages. The first stage Decision base Median Filter (DMF) acts as the preliminary noise removal algorithm. The second stage is either Modified Decision Base Partial Trimmed Global Mean Filter (MDBPTGMF) or Modified Decision Based Unsymmetric Trimmed Median Filter (MDBUTMF) which is used to remove the remaining noise and enhance the image quality. The DMF algorithm performs well at low noise density but it fails to remove the noise at medium and high level. The MDBPTGMF and MDUTMF have excellent performance at low, medium and high noise density but these reduce the image quality and blur the image at high noise level. So the basic idea behind this paper is to combine the advantages of the filters used in both the stages to remove the Salt and Pepper noise and enhance the image quality at all the noise density level. The proposed method is tested against different gray scale images and it gives better Mean Absolute Error (MAE), Peak Signal to Noise Ratio (PSNR) and Image Enhancement Factor (IEF) than the Adaptive Median Filter (AMF), Decision Base Unsymmetric Trimmed Median Filter (DBUTMF), Modified Decision Base Unsymmetric Trimmed Median Filter (MDBUTMF) and Decision Base Partial Trimmed Global Mean Filter (DBPTGMF).

*Keywords*—Salt and Pepper noise, Unsymmetric trimmed Mean Filter, Unsymmetric Trimmed Median Filter

I. INTRODUCTION

The image restoration has a very important role in Digital Image Processing. The digital images can be corrupted by Impulse noise which arises in the image due to bit error in transmission or introduced during the signal acquisition stage. The impulse noise is of two types, one is fixed value impulse noise and other one is random value noise. Salt and pepper noise is a fixed value impulse noise. It can corrupt the image where the corrupted pixel takes either maximum or minimum gray level i.e. 255 or 0. Several non-linear filters have been proposed to remove the salt and pepper noise from the image. Among these standard median filter has been a reliable method to remove salt and pepper noise without damaging the edges. The main disadvantage of standard Median Filter (MF) [1] is that it is only effective in low noise densities. Adaptive Median Filter (AMF) [2] is also performing well at low noise density but at high noise density it blurs the image.

To overcome these drawbacks, different new algorithms have been proposed. Decision Base Algorithm (DBA) [3] is one of them but its disadvantage is that streaking occurs at higher noise densities due to replacement with the neighborhood pixel values. To overcome this drawback, Decision Base Unsymmetric Median Filter (DBUTMF) is proposed [4].Further, algorithms such as Modified Decision Based Unsymmetric Trimmed Median Filter (MDUTMF) [5] and Decision Based Partial Trimmed Global Mean Filter (DBPTGMF) [6] were proposed. These methods give excellent result at low and medium noise densities. But the major disadvantage of these algorithms is that at high noise density i.e. above 60% noise it has a poor IEF value and lead to blurring the image. Previously Cascading Algorithms namely DMF+UTMF and DMF+UTMP [7] Where Decision base Median Filter (DMF) is used in first stage and the Unsymmetric Trimmed Filter (UTF) is used in second stage. Again the UTF may be either Unsymmetric Trimmed Mean Filter (UTMF) or Unsymmetric Trimmed Mid Point filter (UTMP).The first and second stages are connected in cascade. The denoising capability of these cascading algorithms also fails at high noise density.

To overcome this drawback new Cascading algorithms are proposed in this paper. These cascading algorithms efficiently remove the noise at low, medium and high noise level. These also enhance the image quality at high noise level very effectively as compared to the all previous algorithms. These algorithms consist of two stages. First is Decision base Median Filter (DMF) [7] which detects and replaces the corrupted pixels with median value while uncorrupted pixels are left unchanged. The first stage is used to reduce the noise at all noise level. The second stage is one of the two algorithms i.e. either Modified Decision Base Partial Trimmed Global Mean Filter (MDBPTGMF) or Modified Decision Base Unsymmetric Trimmed Median Filter (MDBUTMF). The main objective of the second stage is to remove the remaining noise and increase the image quality at all noise level. These two stages are connected in cascade form. Proposed algorithm PA1 (DMF+MDBPTGMF) produces a consistent result from low to high noise level and it outperforms the previous algorithms at high noise density and Proposed algorithm PA2 (DMF+MDBUTMF) produces excellent result from low to high noise level in terms of PSNR, MAE and IEF as compared to all other algorithms.

The rest of the paper is structured as follows. The proposed algorithms and its different cases are described in section II. Simulation results are presented in section III. Finally conclusions are drawn in section IV.

## II. PROPOSED ALGORITHMS

### A. Stage-1 (*Decision based Median Filter*)

The Decision base Median Filter (DMF) works in the first stage of the proposed algorithm. At first a 3 x 3 window is selected. It decides whether the central pixel is corrupted or not. If it is an uncorrupted pixel, it remains unchanged and if corrupted it is replaced by the median value of the selected window. The algorithm for DMF is described as follows.

*Algorithm:*

Step 1: Select a 2-D window of size 3 x 3. The processing pixel is assumed as $P_{ij}$ which lies at the center of window.

Step 2: If $0 < P_{ij} < 255$, then $P_{ij}$ is considered as uncorrupted pixel and is left unchanged.

Step 3: Otherwise, calculate the median of the pixels in the window and replace processing pixel by the median value.

Step 4: Move the 3 x 3 window to the next pixel in the image. And repeat steps 1 to 3 until all the pixels in the entire image are processed.

The output obtained from DMF is given to the second stage for further processing.

### B. Stage-2

The stage-2 of the proposed cascading algorithm is one of the two algorithms given below.

#### i. *Modified Decision Based Partial Trimmed Global Mean Filter* (*MDBPTGMF*)

The Modified Decision Based Partial Trimmed Global Mean Filter or MDBPTGMF algorithm is a variation to Decision Base Partial Trimmed Global Mean Filter or DBPTGMF algorithm. The major drawback of DBPTGMF algorithm [6], is when a selected window contains only 0 and 255 value then the restored value is either 0 or 255(again noisy), leads us to propose MDBPTGMF. In this algorithm when a selected window contain both the 0 and 255 values then the processing pixel is replaced by mean value of the selected window. The detail of the algorithm is given below.

*Algorithm:*

Step 1: Select a 3 x 3 2-D window. Assume that the processing pixel is $P_{ij}$, which lies at the center of window.

Step 2: If $0 < P_{ij} < 255$, then the processing pixel or $P_{ij}$ is uncorrupted and left unchanged.

Step 3: If $P_{ij} = 0$ or $P_{ij} = 255$, then it is considered as corrupted pixel and four cases are possible as given below.
**Case i):** If the selected window has all the pixel value as 0, then $P_{ij}$ is replaced by the Salt noise(i.e. 255).
**Case ii):** If the selected window contains all the pixel value as 255, then $P_{ij}$ is replaced by the pepper noise (i.e. 0).
**Case iii):** If the selected window contains all the value as 0 and 255 both. Then the processing pixel is replaced by mean value of the window.
**Case iv):** If the selected window contain not all the element 0 and 255. Then eliminate 0 and 255 and find the median value of the remaining element. Replace $P_{ij}$ with median value.

Step 4: Repeat step 1 to 3 for the entire image until the process is complete.

#### ii. *Modified Decision Base Unsymmetric Trimmed Median Filter* (*MDBUTMF*)

In the Modified Decision Base Unsymmetric Trimmed Median filter (MDBUTMF) the pixels are checked to know whether they are noisy or noise free. If the processing pixel is 0 or 255 then it is considered as noisy and it is processed as per the algorithm given below.

*Algorithm:*

Step 1: Select 2-D window of size 3 x 3. Assume that the processing pixel as $P_{ij}$ which lies at the center of window.

Step 2: If $0 < P_{ij} < 255$, then $P_{ij}$ is an uncorrupted pixel and It is left unchanged.

Step 3: If $P_{ij} = 0$ or $P_{ij} = 255$ then it is corrupted pixel and two cases are possible as follow.

**Case i):** If the selected window contains all the elements as 0 and 255, then replace *Pij* with the mean value of elements of the window.

**Case ii):** If the selected window contains not all the element as 0 and 255, then eliminate 0 and 255 and calculate median value of remaining element. Then replace *Pij* with the median value.

Step 4: Repeat steps 1 to 3 until the process is complete for the entire image.

## C. Cascade Filter

The Decision base Median Filter (DMF) is efficient to remove the noise at low noise level only. But at medium and high density it fails because the restored pixel which is the median value of the selected window is also a corrupted pixel value. The MDBPTGMF and MDBUTMF are very efficient to remove the noise but at high noise level it blurs the image which decreases the image quality. Therefore DMF is used to reduce the noise and the MDBPTGMF and MDBUTMF is used to remove the noise completely as well as to enhance the image. In this structure DMF is cascaded with MDBPTGMF or MDBUTMF for the improvement of the output obtained from DMF. The noisy image is first given to stage-1(i.e. to DMF). Then the output of the stage-1 is given as input of stage-2(i.e. either to MDBPTGMF or MDBUTMF). The proposed algorithm-1 (PA1) is the cascaded version of DMF and MDBPTGMF where as proposed algorithm-2 is the cascaded version of DMF and MDBUTMF. The performance of PA2 is better than the performance of PA1. PA1 gives a better result at high noise density where as PA2 is giving an excellent result for all noise level in terms of MAE, PSNR and IEF values as compared to other algorithms.

## III. SIMULATION RESULTS

The proposed algorithms are tested using 512x512 8-bit/pixel image Lena.gif. In the simulation, images are corrupted by Salt and Pepper noise. The noise level varies from 10% to 90% with increment of 10% and the performance is quantitatively measured by Mean Absolute Error (MAE), Peak Signal to Noise Ratio (PSNR) and Image Enhancement Factor (IEF).

$$\text{MAE} = \frac{\sum_{i,j}|Y(i,j)-\hat{Y}(i,j)|}{MXN} \quad (1)$$

$$\text{MSE} = \frac{\sum_{i,j}(Y(i,j)-\hat{Y}(i,j))^2}{MXN} \quad (2)$$

$$\text{PSNR in dB} = 10\log_{10}\left(\frac{255^2}{MSE}\right) \quad (3)$$

$$\text{IEF} = \frac{\sum_{i,j}(\eta(i,j)-Y(i,j))^2}{\sum_{i,j}(\hat{Y}(i,j)-Y(i,j))^2} \quad (4)$$

where, MAE stands for Mean Absolute Error, MSE stands for Mean Square Error, PSNR stands for Peak Signal to Noise Ratio, IEF stands for Image Enhancement Factor. M X N is the size of the image, Y represents original image. Ŷ represents restored image and η represents noisy image. MAE, PSNR and IEF value is calculated for the proposed algorithms and comparison of performance with various filters namely AMF, DBUTMF, MDBUTMF and DBPTGMF are shown in Table I-III. In Table IV-VI the comparison between the proposed algorithms against the existing cascade algorithm (i.e. DMF+UTMF and DMF+UTMP) in terms of MAE, PSNR and IEF values are shown.

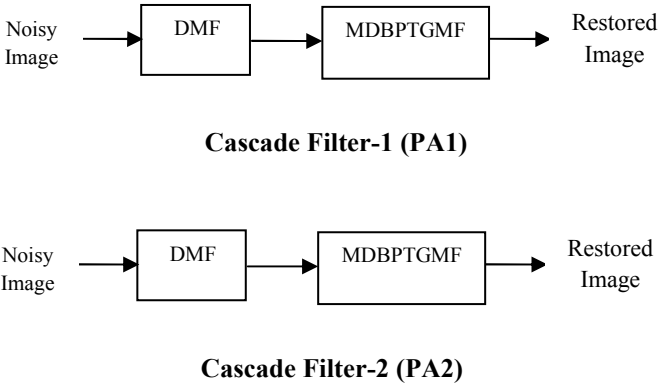

Fig. 1.Cascading process of PA1 and PA2

**Cascade Filter-1 (PA1)**

**Cascade Filter-2 (PA2)**

TABLE-I
COMPARISION OF MAE VALUES OF DIFFERENT ALGORITHMS FOR LENA IMAGE AT DIFFERENT NOISE DENSITIES

| Noise Density in % | MAE VALUES | | | | | |
|---|---|---|---|---|---|---|
| | AMF | DBUTMF | MDBUTMF | DBPTGMF | PA1 | PA2 |
| 10 | 4.99 | 1.57 | 1.01 | 1.62 | 1.97 | 0.17 |
| 20 | 5.53 | 1.73 | 1.54 | 1.81 | 2.01 | 0.36 |
| 30 | 5.85 | 1.96 | 1.83 | 2.04 | 2.12 | 0.62 |
| 40 | 6.10 | 2.37 | 2.22 | 2.43 | 2.16 | 0.92 |
| 50 | 6.49 | 3.48 | 3.12 | 3.22 | 2.48 | 1.33 |
| 60 | 6.71 | 6.32 | 5.90 | 4.90 | 2.97 | 1.89 |
| 70 | 7.37 | 13.92 | 12.71 | 8.18 | 3.80 | 2.80 |
| 80 | 8.59 | 29.55 | 21.76 | 14.08 | 5.78 | 4.61 |
| 90 | 11.50 | 57.01 | 47.98 | 24.09 | 14.74 | 13.21 |

TABLE-II
COMPARISION OF PSNR VALUES OF DIFFERENT ALGORITHMS FOR LENA IMAGE AT DIFFERENT NOISE DENSITIES

| Noise Density in % | PSNR in dB | | | | | |
|---|---|---|---|---|---|---|
| | AMF | DBUT MF | MDBUT MF | DBPTG MF | PA1 | PA2 |
| 10 | 28.39 | 38.20 | 39.95 | 38.08 | 37.19 | 47.71 |
| 20 | 27.55 | 37.57 | 38.54 | 37.47 | 36.91 | 44.54 |
| 30 | 27.09 | 36.91 | 37.33 | 36.78 | 36.65 | 42.01 |
| 40 | 26.71 | 36.06 | 36.65 | 36.01 | 36.34 | 40.21 |
| 50 | 25.90 | 34.37 | 34.92 | 35.01 | 35.82 | 39.32 |
| 60 | 25.75 | 32.60 | 32.94 | 33.10 | 34.73 | 37.02 |
| 70 | 24.69 | 29.85 | 30.77 | 32.79 | 33.65 | 35.20 |
| 80 | 23.22 | 27.21 | 28.19 | 30.96 | 31.80 | 33.64 |
| 90 | 20.55 | 25.08 | 26.09 | 27.35 | 29.32 | 30.95 |

TABLE-III
COMPARISION OF IEF VALUES OF DIFFERENT ALGORITHMS FOR LENA IMAGE AT DIFFERENT NOISE DENSITIES

| Noise Density in % | IEF VALUES | | | | | |
|---|---|---|---|---|---|---|
| | AMF | DBUT MF | MDBUT MF | DBPTG MF | PA1 | PA2 |
| 10 | 24.7 | 417.2 | 594.9 | 511.1 | 277.8 | 682.1 |
| 20 | 33.4 | 361.9 | 444.5 | 395.7 | 307.6 | 568.4 |
| 30 | 47.8 | 382.1 | 465.1 | 368.6 | 376.7 | 494.1 |
| 40 | 58.8 | 271.3 | 323.5 | 313.3 | 321.1 | 388.3 |
| 50 | 67.1 | 126.1 | 287.5 | 272.4 | 291.7 | 329.1 |
| 60 | 43.1 | 86.6 | 170.5 | 197.9 | 201.4 | 290.2 |
| 70 | 28.1 | 44.3 | 98.6 | 115.9 | 165.8 | 198.6 |
| 80 | 7.2 | 19.3 | 56.7 | 84.7 | 90.6 | 117.1 |
| 90 | 1.8 | 5.2 | 12.9 | 18.1 | 38.9 | 41.2 |

From the simulation result shown in Table I to III, it is observed that the performance of proposed algorithm PA1 is improved than the existing algorithms at medium and high noise level (i.e. above 30%) whereas the performance of proposed algorithm PA2 is much improved than the existing algorithms at all noise levels.

TABLE-IV
COMPARISION OF MAE VALUES OF DIFFERENT CASCADE ALGORITHMS FOR LENA IMAGE AT DIFFERENT NOISE DENSITIES

| Noise Density in % | MAE VALUES | | | |
|---|---|---|---|---|
| | DMF+UTMF | DMF+UTMP | PA1 | PA2 |
| 10 | 0.39 | 0.40 | 1.97 | 0.17 |
| 20 | 0.87 | 0.88 | 2.01 | 0.36 |
| 30 | 1.41 | 1.41 | 2.12 | 0.62 |
| 40 | 2.08 | 2.09 | 2.16 | 0.92 |
| 50 | 2.87 | 2.90 | 2.48 | 1.33 |
| 60 | 3.95 | 3.93 | 2.97 | 1.89 |
| 70 | 5.33 | 5.29 | 3.80 | 2.80 |
| 80 | 7.22 | 7.19 | 5.78 | 4.61 |
| 90 | 15.41 | 15.11 | 14.74 | 13.21 |

From Table IV and V, it is clear that the performance of the proposed algorithm PA1 is better than the existing DMF+UTMF and DMF+UTMP cascade algorithms at medium and high frequencies where as the performance of proposed algorithm PA2 is much better than these two existing cascade algorithms at all noise level. Fig 1 shows the plot of MAE against noise densities for Lena image, Fig 2 shows the plot of PSNR against noise densities and Fig 3 shows the plot of IEF against noise densities. Fig 4 and Fig 5 show the reconstructed images using the existing and proposed algorithms on Lena and Boat images corrupted with 70% noise density. From the figure it is clear that the reconstructed images obtained by the proposed algorithms are better than the other existing algorithms.

TABLE-V
COMPARISION OF PSNR VALUES OF DIFFERENT CASCADE ALGORITHMS FOR LENA IMAGE AT DIFFERENT NOISE DENSITIES

| Noise Density in % | PSNR VALUES | | | |
|---|---|---|---|---|
| | DMF+UTMF | DMF+UTMP | PA1 | PA2 |
| 10 | 45.37 | 45.57 | 37.19 | 47.71 |
| 20 | 42.26 | 42.37 | 36.91 | 44.54 |
| 30 | 39.59 | 39.57 | 36.65 | 42.01 |
| 40 | 37.34 | 37.09 | 36.34 | 40.21 |
| 50 | 35.12 | 34.95 | 35.82 | 39.32 |
| 60 | 33.04 | 33.02 | 34.73 | 37.02 |
| 70 | 31.05 | 31.05 | 33.65 | 35.20 |
| 80 | 28.90 | 28.97 | 31.80 | 33.64 |
| 90 | 26.51 | 26.70 | 29.32 | 30.95 |

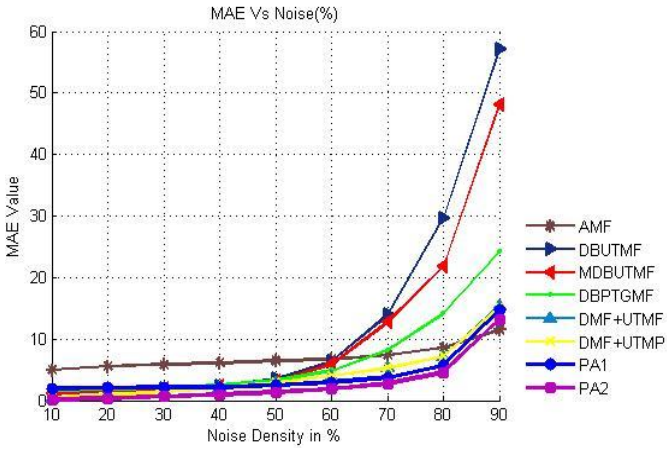

Fig.2.Comparison graph of MAE at different noise densities for 'Lena' Image

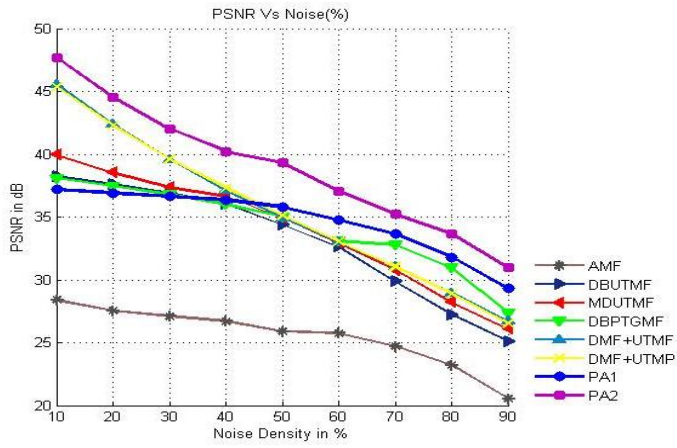

Fig. 3.Comparison graph of PSNR at different noise densities for 'Lena' Image

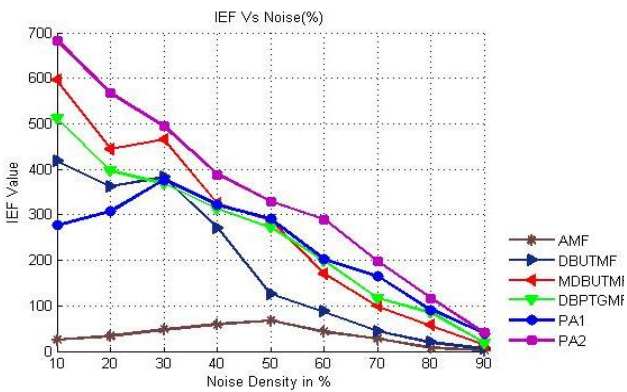

Fig. 4.Comparison graph of IEF at different noise densities for 'Lena' Image

From Fig.2 to 4, it is clear that the proposed algorithm PA1 has better performance than other algorithm at medium and high noise level whereas PA2 has better performance at all noise level.

In Fig 5 and 6, the original, noisy Lena and Boat images are shown and also the reconstructed Lena and Boat images obtained by various filters are given which clearly shows that our proposed methods are better than the other existing algorithm.

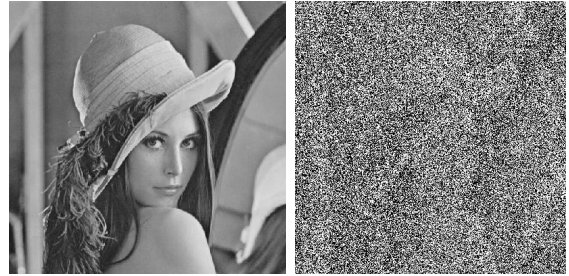

**a**          **b**

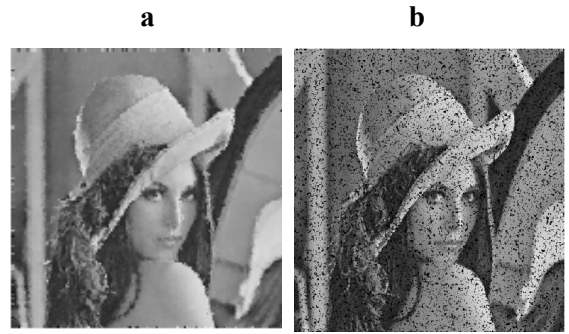

**c**          **d**

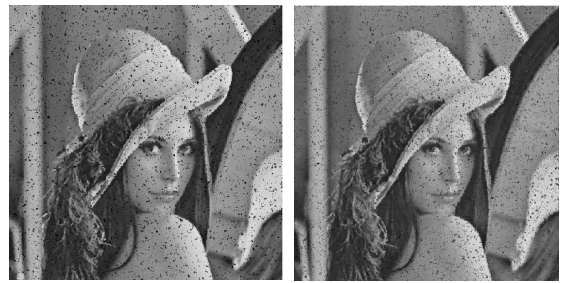

**e**          **f**

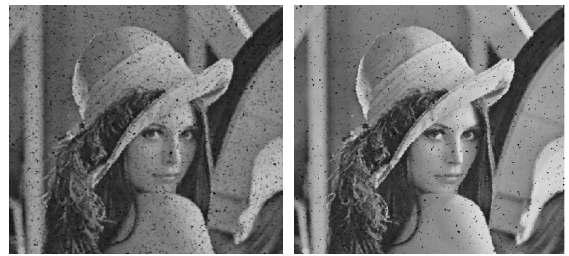

**g**          **h**

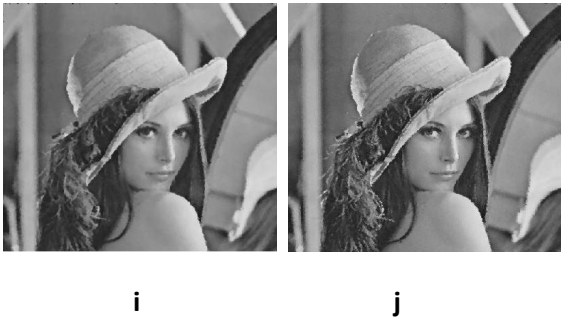
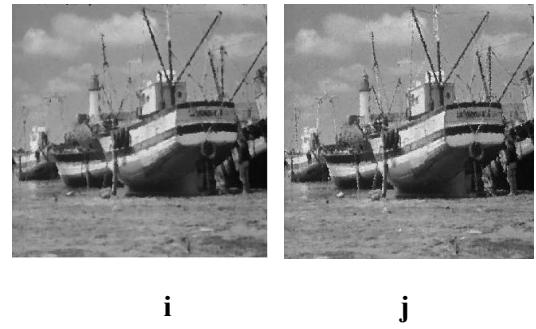

| i | j |

Fig.5.Performance of various filters for Lena image. (a) Original image. (b) Corrupted image with 70% salt and pepper noise. (c) AMF. (d) DBUTMF. (e) MDBUTMF. (f) DBPTGMF. (g) DMF+UTMF. (h) DMF+UTMP. (i) PA1. (j) PA2

Fig.6. Performance of various filters for Boat image. (a) Original image. (b) Corrupted image with 70% salt and pepper noise. (c) AMF. (d) DBUTMF. (e) MDBUTMF. (f) DBPTGMF. (g) DMF+UTMF. (h) DMF+UTMP. (i) PA1. (j) PA2

## IV. CONCLUSION

In this paper, it can be observed that the proposed filters give better result as compared to the existing algorithms in terms of MAE, PSNR and IEF. The proposed algorithms show excellent denoising capability and also preserve texture detail and edges effectively even at very high noise density. The proposed algorithms are effective for removal of salt and pepper noise at low, medium and high noise densities.

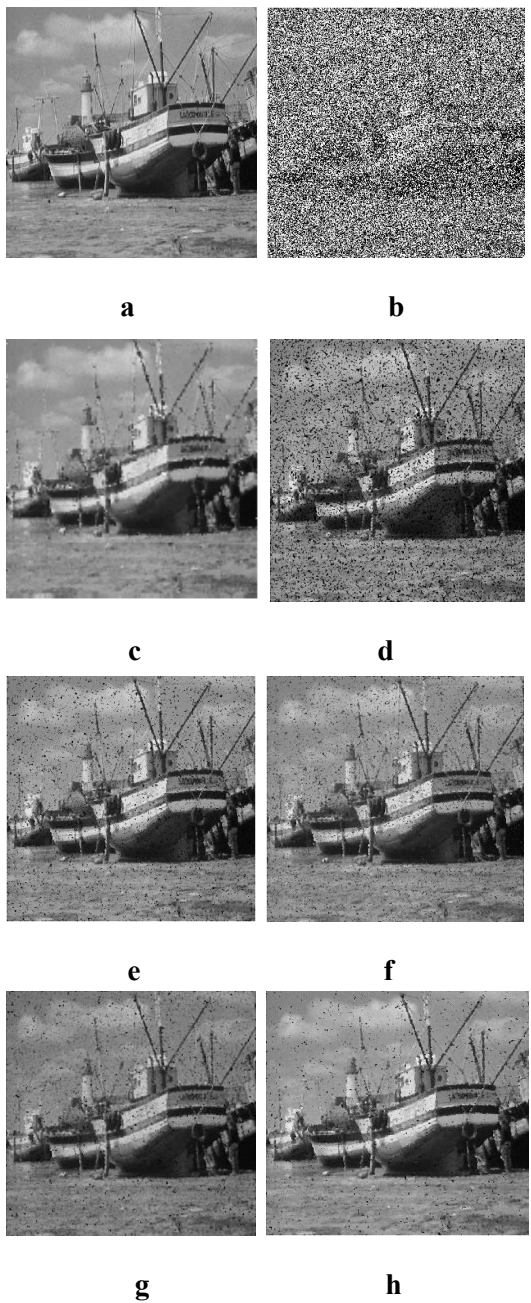

a b
c d
e f
g h

## *References*

[1] J. Astola and P. Kuosmaneen, *Fundamentals of Nonlinear Digital Filtering,* Boca Raton, FL: CRC, 1997.

[2] H. Hwang and R. A. Hadded, " Adaptive median filter: New algorithms and results," *IEEE Trans. Image Process.*, vol. 4, pp. 499-502, Apr. 1995.

[3] K. S. Srinivasan and D. Ebenezer, "A new fast and efficient decision based algorithm for removal of high density impulse noise," *IEEE Signal Process. Lett.*, vol. 14, no. 6, pp. 1506-1516, Jun. 2006.

[4] K. Aiswarya, V. Jayaraj, and D. Ebenezer, " A new and efficient algorithm for the removal of high density salt and pepper noise in images and videos," in *Second Int. Conf. Computer Modeling and Simulation,*2010, pp. 409-413

[5] S. Esakkirajan, T. Veerakumar, A. N. Subramanyam, and C. H. PremChand, " Removal of high density salt and pepper noise through modified decision base unsymmetric trimmed median filter," *IEEE Signal process. Lett,*vol. 18, pp. 287-290, May. 2011.

[6] M.T.Raza,And S. Sawant, "High density salt and pepper noise removal through decision based partial trimmed global mean filter," *Engineering (NUiCONE), 2012 Nirma University International Conference on,* pp.1-5, IEEE, Dec. 2012

[7] S. Balasubramanian, S. Kalishwaran, R. Muthuraj, D. Ebenezer, and V. Jayaraj. "An efficient non-linear cascade filtering algorithm for removal of high density salt and pepper noise in image and video sequence." In *Control, Automation, Communication and Energy Conservation, 2009. INCACEC 2009,* pp. 1-6.,IEEE, Jun. 2009.